\documentclass[conference]{IEEEtran}
\IEEEoverridecommandlockouts
\usepackage{amsmath, amsthm, amsfonts,amssymb,euscript, graphicx,epsfig,enumerate,float,afterpage, subfigure, ifthen, moreverb, algpseudocode,algorithm}
\usepackage{hyperref}
\usepackage{tikz}
\usepackage{placeins}
\usepackage{subfigure}
\usetikzlibrary{arrows.meta, positioning, calc}
\usepackage{xcolor}

\newtheorem{proposition}{Proposition}

\newcommand{\curl}[1]{\left\{#1\right\}}
\newcommand{\vect}[1]{\mathbf{#1}}
\newcommand{\parn}[1]{\left(#1\right)}

\newcommand{\sqr}[1]{\left[#1\right]}

\begin{document}

\title{\huge{Causal Discovery in Equal Variance Linear Gaussian DAGs via SURE-Tuned Ridge Regression}}

\author{\IEEEauthorblockN{Sambit Mishra \& Urbashi Mitra}
\IEEEauthorblockA{\textit{Department of Electrical and Computer Engineering} \\
\textit{University of Southern California} \\
\{sambitmi, ubli\}@usc.edu}\thanks{This work has been funded by one or all of the following grants: ARO W911NF1910269, ARO W911NF2410094, ONR N00014-22-1-2363, NSF CIF-2311653, NSF CIF-2148313, NSF RINGS-2148313, NSF DBI-2412522, and is also supported in part by funds from federal agency and industry partners as specified in the RINGS program. }
}

\maketitle

\begin{abstract}
Recovering the directed acyclic graph (DAG) of a structural equation model (SEM) from observational data is a central problem in causal discovery. The iterative gradient descent and per-problem hyperparameter tuning of continuous-optimization methods are poorly suited to two practically important regimes: the sample-limited regime, where the number of samples is comparable to or smaller than the number of nodes in the DAG, and the compute-limited regime. This work proposes SURE-Ridge, a non-iterative, closed-form estimator for equal variance linear Gaussian SEM. The method performs parallel node-wise regressions with regularization parameters chosen adaptively by Stein's unbiased risk estimate (SURE), and applies an adaptive thresholding procedure to extract a DAG from the resulting soft adjacency matrix. Numerical results show that SURE-Ridge achieves the lowest structural Hamming distance in the small-sample regime and the lowest run time across all sample sizes tested, compared with NOTEARS, DAGMA, and GBNSL baselines.
\end{abstract}


\section{Introduction}
Linear structural equation models (SEMs) provide a flexible framework for representing causal relationships among random variables, and recovering the underlying directed acyclic graph (DAG) from the observational data is a central problem in causal discovery, with applications spanning epidemiology \cite{rothman2005causation}, wireless networks \cite{mata2025integrating}, and financial fraud detection \cite{vivek2024explainable}. 
For linear Gaussian SEMs with equal noise variance across all nodes, the true DAG is identifiable from the joint distribution alone \cite{peters2014identifiability}.  We consider two key regimes herein:  (a) sample-limited regimes (as considered in \cite{shaska2025causal,valen_isit26}), where the number of samples is often comparable to or even smaller than the number of nodes, as in \cite{nguyen2021comprehensive,zhang2020causal}, and (b) compute-limited regimes where a DAG must be continuously estimated as in online and streaming applications \cite{yu2012exploring}.

Classical approaches such as the PC \cite{spirtes1991algorithm} and GES \cite{chickering2002optimal} have computational costs that scale poorly with the number of nodes in the DAG. Continuous optimization formulations introduced by NOTEARS \cite{zheng2018dags} and DAGMA \cite{bello2022dagma} recast structure learning as a smooth, constrained optimization problem via a differentiable acyclicity functional, but they require thousands of gradient steps and per-problem hyperparameter tuning, and are thus not well-matched to the two regimes under consideration.  

Work that directly exploits the equal-variance assumption directly considers two principles. The first is iterative peeling from the inverse covariance matrix: GBNSL in \cite{ghoshal2017learning} identifies terminal nodes one at a time, and BUILD \cite{ajorlou2025build} prunes leaves bottom-up. The second approach is staged regression, in which ordering and parent selection are separated into sequential phases: first ranking nodes by conditional variances and then fitting parents on the recovered order \cite{chen2019causal}. Both strategies treat ordering recovery as an explicit intermediate step, in contrast to SURE-Ridge,
%, sequential in the iterative case and staged in the regression case, 
and both compound finite-sample errors across stages when the sample size is comparable to the number of nodes in the DAG. 

We propose {\bf SURE-Ridge}, a non-iterative, closed-form estimator for equal-variance linear Gaussian DAG recovery. The method performs parallel node-wise ridge regressions, with the $L_2$ regularization parameter chosen by minimizing Stein's unbiased risk estimate (SURE) \cite{stein1981estimation}. SURE provides a tractable estimate of the prediction error 
%of a linear smoother, 
yielding a tuning rule that requires neither held-out data nor iterative search, thus providing a decisive advantage over cross-validation when samples are scarce and over gradient-based tuning when many DAGs must be estimated. Pairing SURE with ridge regression is natural under the equal-variance assumption, where each node's structural equation is itself a linear Gaussian regression on its parents, and subproblems decouple, enabling parallel discovery.
%with a shared noise scale
%, leading to node-wise regressions being a natural fit, the noise level needed by SURE is shared across nodes, and the subproblems decouple and can be solved in parallel without any sequential ordering recovery.
The contributions of this paper are as follows:
\begin{enumerate}
    \item A closed-form formulation of equal variance linear Gaussian DAG recovery via SURE-tuned node-wise regression with total computational cost $\mathcal{O}\parn{d^4 + d^3 n + dM\parn{d - 1} + d^3\log d}$ per DAG;
    \item An adaptive thresholding procedure that requires only two scale-invariant constants and produces a valid DAG in the typical case;
    \item Numerical results demonstrating that SURE-Ridge achieves the lowest structural Hamming distance (SHD) in the small-sample regime and the lowest run time across all sample sizes tested, against NOTEARS, DAGMA, and GBNSL baselines.
\end{enumerate}

This paper is organized as follows. Section \ref{sec:sys} describes the system model and notation. Section \ref{sec:sure} derives the SURE-tuned ridge estimator. 
%and its fast evaluation via eigendecomposition.
Section \ref{sec:thres} presents the adaptive thresholding procedure. Section \ref{sec:experiments} reports numerical results and Section \ref{sec:conclusion} concludes the work.

\section{System Model}
\label{sec:sys}
We consider a $d-$node causal DAG $\mathcal{G} = \parn{\mathcal{V}, \mathcal{E}}$, where $\mathcal{V} = \curl{1, \dots, d}$ represents the node set and $\mathcal{E} = \curl{\parn{i,j}: \text{edge } i \rightarrow j \text{ exists}}$ represents the edge set.  We represent the graph using a weighted adjacency matrix $\vect{W} \in \mathbb{R}^{d \times d}$ under the convention
\begin{equation}
    \vect{W}_{i,j} \neq 0 \iff \parn{j,i} \in \mathcal{E}.
\end{equation}
For each node $i \in \mathcal{V}$, we define the row-vector of causal coefficients, $\vect{w}_i \in \mathbb{R}^{1 \times \parn{d - 1}}$,  ($i^{th}$ row of $\vect{W}$ with $\vect{W}_{i,i}$ removed),
\begin{equation}
    \vect{w}_i \triangleq \vect{W}_{i, -i}.
\end{equation}
Let $X_i$ be the random variable associated with node $i$ and $\vect{x}_i \in \mathbb{R}^{1 \times n}$ be the row-vector of $n$ i.i.d. observations of $X_i$. The concatenation of observations is denoted $\vect{X} \in \mathbb{R}^{d \times n}$, where $\vect{X}_{i, :} = \vect{x}_i$. Under a linear Gaussian SEM with equal noise variance,
\begin{equation}
    \vect{x}_i = \vect{w}_i\vect{X}_{-i, :} + \vect{n}_i,
    \label{eq3}
\end{equation}
where $\vect{X}_{-i, :}$ represents the data matrix $\vect{X}$ with the $i^{th}$ row removed, and $\vect{n}_i \sim \mathcal{N}\parn{\vect{0}, \sigma^2\vect{I}_n}$ and i.i.d. with respect to $i$. This case is identifiable as per \cite{peters2014identifiability}.

\section{SURE-Tuned Ridge Regression}
\label{sec:sure}
We see that \eqref{eq3} can be interpreted as a regression to determine $X_i$ from noisy observations governed by the causal coefficient vector $\vect{w}_i$. Thus, 
we can estimate $\vect{w}_i$ using ridge regression, with the tunable regularization parameter $\lambda_i$. Let the Gram matrix corresponding to node $i$ be denoted as $\vect{G}_i = \vect{X}_{-i, :}\vect{X}_{-i, :}^T$, then we have the ridge regression-based estimator of $\vect{w}_i$ given by
\begin{equation}
    \hat{\vect{w}}_{i}\parn{\lambda_i} = \parn{\vect{G}_i + \lambda_i\vect{I}_{d - 1}}^{-1}\vect{X}_{-i,:}\vect{x}_i^T.
    \label{eq4}
\end{equation}
A principled choice of the tunable parameter $\lambda_i$ is critical, as it directly controls the bias-variance trade-off of the resulting estimator $\hat{\vect{w}}_{i}\parn{\lambda_i}$. We adopt SURE as a data-driven criterion for selecting $\lambda_i$. The following proposition formalizes the per-node SURE objective.
\begin{proposition}[Per-node SURE Objective]
\label{prop:sure}
Let the noise vector satisfy $\vect{n}_i^T \sim \mathcal{N}\parn{\vect{0}, \sigma^2\vect{I}_n}$, and let $\hat{\vect{w}}_i\parn{\lambda_i}$ be the ridge estimator defined in \eqref{eq4}. Then an unbiased estimator of the expected squared-error risk $\mathbb{E}\sqr{\|\vect{x}_i^T - \vect{X}_{-i,:}^T\hat{\vect{w}}_i^T\parn{\lambda_i}\|_2^2} - n\sigma^2$ is given by
\begin{eqnarray}
    \operatorname{SURE}_i\parn{\lambda_i} &=& -n\sigma^2 \nonumber \\
    && \!\!\!\!\!\! + \|\vect{x}_i^T - \vect{X}_{-i,:}^T\parn{\vect{G}_i + \lambda_i\vect{I}_{d - 1}}^{-1}\vect{X}_{-i, :}\vect{x}_i^T\|_2^2 \nonumber \\
    && \!\!\!\!\!\! + 2\sigma^2\operatorname{tr}\sqr{\parn{\vect{G}_i + \lambda_i\vect{I}_{d - 1}}^{-1}\vect{G}_i}.
    \label{eq5}
\end{eqnarray}
\end{proposition}

The proposition follows from applying the SURE construction in \cite{stein1981estimation} to the regression model in \eqref{eq3}. The objective in \eqref{eq5} thus gives us a principled risk metric for choosing the optimal parameter $\lambda_i$ for each node $i \in \mathcal{V}$, which we obtain by solving the following optimization problem
\begin{equation}
    \lambda_i^* = \arg\min_{\lambda_i > 0} \operatorname{SURE}_i\parn{\lambda_i}.
\end{equation}

%A direct grid search over $\lambda_i$ would require recomputing $\parn{\vect{G}_i + \lambda_i\vect{I}_{d - 1}}^{-1}$ at every grid point, but a single 
We further massage the objective function via the 
eigendecomposition $\vect{G}_i = \vect{U}_i\boldsymbol{\Gamma}_i\vect{U}_i^T$, where $\vect{U}_i$ is the orthogonal matrix of eigenvectors of $\vect{G}_i$ and $\boldsymbol{\Gamma}_i = \operatorname{diag}\parn{\gamma_{i,1}, \dots, \gamma_{i,d-1}}$ collects the eigenvalues. Defining the projection $\vect{z}_i = \vect{U}_i^T\vect{X}_{-i,:}\vect{x}_i^T$ 
translates the SURE objective to a sum of scalar functions of $\lambda_i$ that can be evaluated at $\mathcal{O}\parn{d - 1}$ cost per grid point \cite{hastie2009elements}. The SURE objective in \eqref{eq5} admits the following closed-form expression
\begin{eqnarray}
    \operatorname{SURE}_i\parn{\lambda_i} \! \! \!  &=&  \! \!\!-n\sigma^2 + \|\vect{x}_i\|_2^2 - 2\sum_{k = 1}^{d - 1}\frac{z_{i,k}^2}{\gamma_{i,k} + \lambda_i} \nonumber \\
    &&  \! \! \!\! \! \! \! \! \!\! \! \!+ \sum_{k = 1}^{d - 1}\frac{\gamma_{i,k}\,z_{i,k}^2}{\parn{\gamma_{i,k} + \lambda_i}^2} + 2\sigma^2\sum_{k = 1}^{d - 1}\frac{\gamma_{i,k}}{\gamma_{i,k} + \lambda_i}. 
    \label{eq7}
\end{eqnarray}
Given the optimal parameter $\lambda_i^*$, the corresponding closed form ridge estimate is
\begin{eqnarray}
    \hat{\vect{w}}_i^* = \vect{z}_i^T\operatorname{diag}\parn{\frac{1}{\gamma_{i,k} + \lambda_i^*}}_{k=1}^{d-1}\vect{U}_i^T.
\end{eqnarray} Finally, concatenating all the $\hat{\vect{w}}_i^*$ row-wise with additional diagonal element set to $0$ leads us to the final soft estimate of the adjacency matrix $\tilde{\vect{W}}_{\operatorname{est}}$. 

%The equal-noise-variance assumption ensures that the true DAG is uniquely identifiable from the joint distribution of $\parn{X_1, \dots, X_d}$ \cite{peters2014identifiability}. 
The SURE-tuned ridge estimator captures the population-level regression structure that ensures identifiability under the noise-equal-variance assumption \cite{peters2014identifiability}, and the adaptive thresholding stage of Section \ref{sec:thres} extracts a DAG estimate from the resulting soft adjacency matrix. 
%A formal finite-sample recovery analysis is beyond the scope of this work, we instead validate the procedure empirically in Section \ref{sec:experiments}.

\section{Adaptive Thresholding}
\label{sec:thres}
The soft adjacency matrix $\tilde{\vect{W}}_{\operatorname{est}}$ is typically not a DAG. To extract a valid DAG structure, we propose a three-stage procedure. We start with an absolute floor threshold that scales with the signal magnitude, then perform a discrete bisection search over the above-floor magnitudes, and lastly consider a fallback that preserves informative edges when the bisection becomes too aggressive.

\subsection{Acyclicity Test}
We use the polynomial acyclicity functional of \cite{yu2019dag},
\begin{eqnarray}
    h\parn{\vect{W}} = \operatorname{tr}\sqr{\parn{\vect{I}_d + \frac{\vect{W} \odot \vect{W}}{d}}^d} - d,
    \label{eq9}
\end{eqnarray}
which is a numerically stable variant of the matrix exponential form originally proposed in \cite{zheng2018dags}. It can be verified that $h\parn{\vect{W}} = 0$ if and only if $\vect{W}$ represents a DAG. Evaluating $h$ at a candidate thresholded graph is $\mathcal{O}\parn{d^3}$ in computational complexity and vectorizes trivially across a batch of candidate matrices. 

\subsection{Signal-Scaled Floor Filter}
A simple threshold of zero retains all the ridge-shrinkage noise, while a bisection starting from zero can, in dense regimes, push the threshold above the true edge magnitudes and return an empty graph. We therefore introduce a scale-adaptive floor
\begin{eqnarray}
    \eta_{\operatorname{min}} = \max\parn{\eta_0, \beta\max_{i \neq j}\left|\parn{\tilde{\vect{W}}_{\operatorname{est}}}_{i,j}\right|},
\end{eqnarray}
where $\eta_0$ and $\beta$ are tunable hyperparameters governed by the number of nodes $d$ and the expected signal-to-noise ratio (SNR). Since the ridge shrinkage is bounded above by OLS,  $\max\left|\tilde{\vect{W}}_{\operatorname{est}}\right|$ tracks the largest signal magnitude. In small-to-moderate-$d$ or high-SNR settings, the $\beta$-scaled term is highly significant and acts as a proper scale-adaptive floor, and therefore it is set high in such settings to aggressively filter out spurious correlations. However, in large-$d$ or low-SNR regimes, heavy ridge regularization severely suppresses the maximum edge weights toward the noise floor, rendering the $\beta$ term insufficient. Consequently, $\eta_0$ acts as an absolute safety bound and is tuned as per the expected noise parameters to separate ambient noise artifacts from the heavily penalized true values.

The first stage then zeroes out all entries below $\eta_{\operatorname{min}}$. Formally, we have
\begin{eqnarray}
    \vect{W}_{\operatorname{floor}} = \mathbb{I}\parn{\left|\tilde{\vect{W}}_{\operatorname{est}}\right| > \eta_{\operatorname{min}}} \odot \tilde{\vect{W}}_{\operatorname{est}}.
\end{eqnarray}
If we obtain $h\parn{\vect{W}_{\operatorname{floor}}} = 0$, in this thresholding step, the procedure stops with $\vect{W}_{\operatorname{floor}}$ as the final estimate.
%then we consider it to be the final estimated DAG.

\subsection{Bisection Over Above-Floor Magnitudes}
When $\vect{W}_{\operatorname{floor}}$ is still cyclic, we search for a larger threshold that restores acyclicity. The candidate set is $\curl{\eta_{\min}} \cup \curl{|\tilde{\vect{W}}_{\operatorname{est}, ij}| : |\tilde{\vect{W}}_{\operatorname{est}, ij}| > \eta_{\min}}$, sorted in ascending order. Because the map $\eta \mapsto \mathbb{I}\sqr{\mathbb{I}\parn{|\tilde{\vect{W}}_{\operatorname{est}}| > \eta}\odot\tilde{\vect{W}}_{\operatorname{est}}\text{ is a DAG}}$ is upward-closed on this sorted list, the smallest acyclic candidate $\eta^*$ can be located by discrete binary search in at most $\lceil2\log_2 d\rceil$ acyclicity checks. We denote the resulting estimate $\vect{W}_{\operatorname{bisect}}$.

\subsection{Fallback for Aggressive Cuts}
Aggressive bisection can remove most floor-thresholded edges to break a small number of cycles, yielding a near-empty DAG that is less informative than a mildly cyclic matrix. To guard against this, we return
\begin{equation}
    \vect{W}_{\operatorname{est}} = \begin{cases} \vect{W}_{\operatorname{floor}} & \text{if } \|\vect{W}_{\operatorname{bisect}}\|_0 < \tfrac{1}{2}\|\vect{W}_{\operatorname{floor}}\|_0, \\ \vect{W}_{\operatorname{bisect}} & \text{otherwise,} \end{cases}
\end{equation}
where $\|\cdot\|_0$ counts the number of nonzero entries. The fallback prioritizes edge preservation over strict acyclicity. $\vect{W}_{\operatorname{est}}$ is a valid DAG whenever the floor or bisection stage succeeds which commonly occurred in our numerical results.
%, which is the common case at the sample sizes and sparsity levels we study; 
In the rare fallback case $\vect{W}_{\operatorname{est}}$ may contain a small number of cycles. 

The entire algorithm is summarized in Algorithm~\ref{alg:sure}. The dominant per-node cost in the SURE-tuned ridge stage is the eigendecomposition of $\vect{G}_i$, which is $\mathcal{O}\parn{\parn{d - 1}^3}$, followed by the projection $\vect{z}_i = \vect{U}_i^T\vect{X}_{-i, :}\vect{x}_i^T$ at $\mathcal{O}\parn{\parn{d - 1}^2 n}$ and the SURE grid evaluation at $\mathcal{O}\parn{M\parn{d - 1}}$. Since these steps are repeated for each of the $d$ nodes, the total cost of the SURE-tuned ridge stage is $\mathcal{O}\parn{d\parn{d - 1}^3 + d\parn{d - 1}^2 n + dM\parn{d - 1}}$. The thresholding stage adds a single $\mathcal{O}\parn{d^3}$ acyclicity evaluation in the floor filter, and at most $\lceil2\log_2 d\rceil$ further $\mathcal{O}\parn{d^3}$ evaluations during the bisection. Therefore, the worst-case overall complexity of Algorithm \ref{alg:sure} is $\mathcal{O}\parn{d^4 + d^3 n + dM\parn{d - 1} + d^3\log d}$ per DAG.

\begin{algorithm}
\caption{SURE-Ridge DAG Recovery}
\label{alg:sure}
\begin{algorithmic}[1]
\Require Data matrix $\vect{X} \in \mathbb{R}^{d \times n}$, noise variance $\sigma^2$, grid $\Lambda$, floor $\eta_0$, scale $\beta$.
\Ensure Estimated adjacency matrix $\vect{W}_{\operatorname{est}} \in \mathbb{R}^{d \times d}$.
\For{$i = 1, \ldots, d$}
    \State Form $\vect{X}_{-i, :}$ and $\vect{G}_i = \vect{X}_{-i, :}\vect{X}_{-i, :}^T$; eigendecompose $\vect{G}_i = \vect{U}_i\boldsymbol{\Gamma}_i\vect{U}_i^T$.
    \State Compute $\vect{z}_i = \vect{U}_i^T\vect{X}_{-i, :}\vect{x}_i^T$ and $\operatorname{SURE}_i\parn{\lambda}$ for each $\lambda \in \Lambda$.
    \State $\lambda_i^* \gets \arg\min_{\lambda \in \Lambda}\operatorname{SURE}_i\parn{\lambda}$; recover $\parn{\hat{\vect{w}}_i^T}^*$ using the eigendecomposition.
\EndFor
\State Assemble $\tilde{\vect{W}}_{\operatorname{est}}$ from $\curl{\parn{\hat{\vect{w}}_i^T}^*}$ with zero diagonal.
\State Compute $\eta_{\min} = \max\curl{\eta_0,\ \beta\max_{i \neq j}|\tilde{\vect{W}}_{\operatorname{est}, ij}|}$ and set $\vect{W}_{\operatorname{floor}} = \mathbb{I}\parn{|\tilde{\vect{W}}_{\operatorname{est}}| > \eta_{\min}} \odot \tilde{\vect{W}}_{\operatorname{est}}$.
\If{$h\parn{\vect{W}_{\operatorname{floor}}} = 0$}
    \State \Return $\vect{W}_{\operatorname{floor}}$.
\EndIf
\State Binary-search the sorted above-floor magnitudes for the smallest $\eta^*$ giving an acyclic $\vect{W}_{\operatorname{bisect}} = \mathbb{I}\parn{|\tilde{\vect{W}}_{\operatorname{est}}| > \eta^*} \odot \tilde{\vect{W}}_{\operatorname{est}}$.
\If{$\|\vect{W}_{\operatorname{bisect}}\|_0 < \tfrac{1}{2}\|\vect{W}_{\operatorname{floor}}\|_0$}
    \State \Return $\vect{W}_{\operatorname{floor}}$.
\Else
    \State \Return $\vect{W}_{\operatorname{bisect}}$.
\EndIf
\end{algorithmic}
\end{algorithm}

\section{Numerical Results and Discussion}
\label{sec:experiments}
\subsection{Experimental Setup}
We simulate Gaussian SEMs with $d \in \curl{20, 50}$ nodes and $\lceil2d/5\rceil$ edges per graph, drawn from a sparse random-DAG distribution. This corresponds to a sub-Erd\H{o}s--R\'{e}nyi-1 average in-degree, matching the sparse regime targeted by equal-variance methods. 
%Performance in denser regimes is left for future investigation. 
Non-zero entries of $\vect{W}$ are sampled uniformly from $\sqr{-2.0, -1.0} \cup \sqr{1.0, 2.0}$. The noise variance is fixed at $\sigma^2 = 1$. 
%For each $d$ we sweep $n$ roughly from $20$ to $200$ in steps of $20$, with 
We have
$100$ Monte Carlo trials per setting (values of $n$ and $d$). Each method observes the same $\parn{\vect{W}_{\operatorname{true}}, \vect{X}}$ per trial to enable fair comparisons, where $\vect{W}_{\operatorname{true}}$ is the ground truth DAG adjacency matrix.

We compare SURE-Ridge against three baselines: NOTEARS, DAGMA, and GBNSL. NOTEARS \cite{zheng2018dags} solves
\begin{eqnarray}
    \min_{\vect{W} \in \mathbb{R}^{d \times d}} \; \tfrac{1}{2n}\|\vect{X} - \vect{WX}\|_F^2 + \lambda \|\vect{W}\|_1 \quad \text{s.t.} \quad h\parn{\vect{W}} = 0,
\end{eqnarray}
with the trace-exponential acyclicity functional $h\parn{\vect{W}} = \operatorname{tr}\parn{e^{\vect{W} \odot \vect{W}}} - d$, via an augmented Lagrangian. We substitute the polynomial form in \eqref{eq9} for numerical stability. DAGMA \cite{bello2022dagma} swaps this constraint for a log-determinant form,
\begin{eqnarray}
    h_{\text{ldet}}\parn{\vect{W};s} = -\log\det\parn{s\vect{I}_{d} - \vect{W} \odot \vect{W}} + d \log s,
\end{eqnarray}
which has better-behaved gradients, and replaces the augmented Lagrangian with a central-path scheme that solves a sequence of unconstrained subproblems. 

GBNSL \cite{ghoshal2017learning} uses successive node pruning to determine the graph and does not require gradient-based optimization. Given a sparse $\ell_1$-regularized estimate $\hat{\boldsymbol{\Omega}}$ of the precision matrix and least-squares estimates $\hat{\boldsymbol{\theta}}^i$ of the regression of each $X_i$ on its Markov blanket (parents and children), GBNSL uses the identity that under equal noise variance $\sigma^2$, node $i$ is a leaf node if and only if
\begin{eqnarray}
    \theta^i_j = \sigma^2\,\Omega_{i,j}, \quad \forall j \neq i.
\end{eqnarray}
At each iteration, leaf nodes are determined as those minimizing $|\hat{\Omega}_{i,j}/\hat{\theta}^i_j|$ over its Markov blanket, after which $\hat{\boldsymbol{\Omega}}$ is updated by a rank-1 Schur complement.  Conditioned on the found leaf nodes enables the next iteration which continues until the causal ordering is determined.  A final least-squares estimation yields the graph edge weights.
%and the procedure repeats until the causal ordering is recovered, with a final OLS pass yielding the edge weights; without equal variance, these ratios no longer share a common value across $j$ and the rule fails. 
NOTEARS and DAGMA are initialized to zero with a matched budget of $32000$ Adam steps (learning rate $3 \times 10^{-4}$, $L_1$ penalty $5 \times 10^{-2}$): NOTEARS uses $32$ outer iterations of $1000$ inner steps, DAGMA uses $T = 4$ central-path stages of $8000$ inner steps with $\mu_0 = 1.0$ and decay $\alpha = 0.1$. GBNSL's $\ell_1$ parameter and Markov-blanket threshold are tuned once and held fixed. All three baselines apply a post-hoc threshold of $0.1$ before scoring. We measure accuracy by normalized SHD,
\begin{eqnarray}
    \text{Normalized SHD} = \frac{2\,\operatorname{SHD}}{d\parn{d - 1}},
\end{eqnarray}
where SHD is the minimum number of edge insertions, deletions, or reversals needed to transform one graph into another. The code for the implementation can be found at \url{https://github.com/SamMathelete/SURE_Ridge_Final}.

\subsection{Sample Efficiency}
\begin{figure}
    \centering
    \includegraphics[width=0.95\columnwidth]{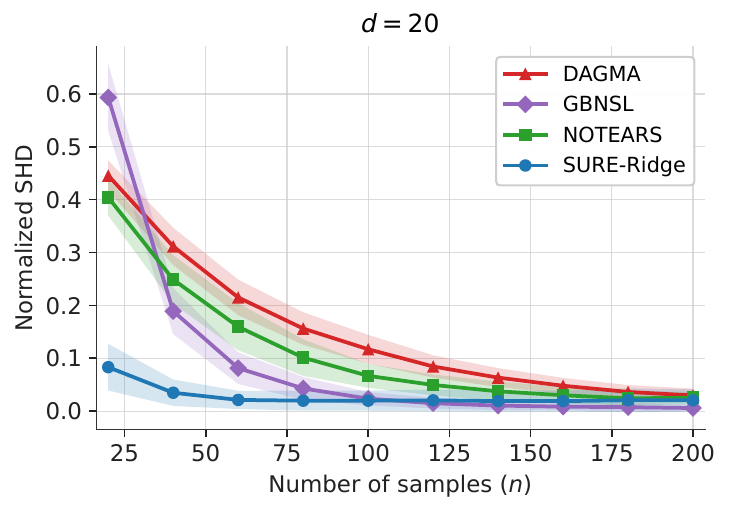}
    \\[-2pt]
    \includegraphics[width=0.95\columnwidth]{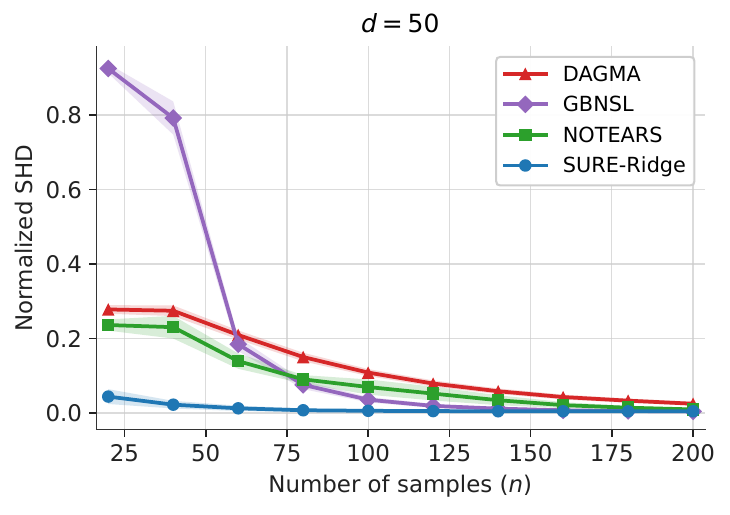}
    \vspace*{-0.1in}
    \caption{SHD vs.\ sample size $n$ across two problem sizes ($d = 20, 50$, top to bottom). Lines and shaded bands show the mean and one standard deviation across 100 Monte Carlo trials per setting.}
    \label{fig:shd}
\end{figure}

Figure \ref{fig:shd} shows the normalized SHD as a function of $n$ at $d = 20$ and  $d = 50$. The four methods exhibit qualitatively distinct sample-efficiency profiles, and the ranking is regime-dependent. In the sample-limited regime ($n$ on the order of $d$ or smaller), SURE-Ridge attains the lowest normalized SHD by a substantial margin, with the gap widening as $d$ grows. This indicates that the advantage is not a small $d$ artifact. The two iterative methods, NOTEARS and DAGMA, perform considerably worse here, and GBNSL performs the worst of the four at very small $n$, with normalized SHD close to $1$ 
%at the smallest sample sizes 
for $d = 50$. 
This is consistent with the high-dimensional regime, where the guarantees of GBNSL's CLIME-based \cite{cai2011constrained} precision estimate do not yet apply. As $n$ grows, the ordering changes: GBNSL improves the fastest of any method and overtakes the iterative baselines near $n \approx 2d,3d$, eventually reaching parity with and slightly surpassing SURE-Ridge near the upper end of the sweep, while SURE-Ridge stays approximately flat at a small, but non-zero value across the entire range, which is consistent with the small residual bias introduced by its post-hoc thresholding step. NOTEARS and DAGMA decrease monotonically, %with NOTEARS consistently slightly better than DAGMA, 
but neither reaches SURE-Ridge or GBNSL.
%within the tested range. 
Thus, SURE-Ridge dominates when samples are scarce, GBNSL dominates when samples are plentiful, and the iterative methods are outperformed throughout the small-sample regime.

\subsection{Run Time Comparison}
\begin{figure}[t]
\centering
\includegraphics[width=0.95\columnwidth]{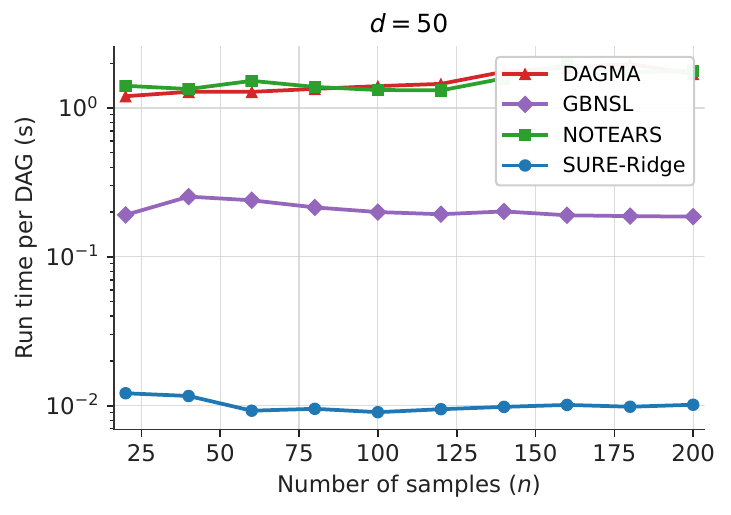}
 \vspace*{-0.1in}
\caption{Per-DAG run time on an Intel i9-13900F CPU at $d = 50$, on a logarithmic scale.}
\label{fig:walltime}
\end{figure}
Figure \ref{fig:walltime} reports per-DAG run time at $d = 50$ on a logarithmic scale, and three distinct tiers are visible.  NOTEARS has slightly improved run time over DAGMA, as the polynomial acyclicity functional is cheaper per gradient step than the log-determinant constraints in DAGMA.
%, which requires a Cholesky-type factorization at every iteration. 
GBNSL sits roughly an order of magnitude below the iterative methods, dominated by the cost of solving linear programs and regression problems.
%$d$ CLIME linear programs and a sequence of OLS regressions on small Markov-blanket sub-problems. 
SURE-Ridge is another magnitude below GBNSL, putting it roughly two orders of magnitude faster than the iterative baselines at this problem size. All four runtimes are essentially flat in $n$.
%, since none of the methods exhibit a leading-order $ n$-dependence in the regime tested. 
%SURE-Ridge's per-node eigendecomposition dominates over its $\mathcal{O}\parn{n}$ projection, GBNSL is dominated by the linear programs which are independent of $n$, and the iterative methods are dominated by their fixed Adam budgets.

\subsection{Discussion}
SURE-Ridge's small-sample dominance and speed advantage stem from two structural features. First, sparse DAGs set most node-wise regression coefficients to zero, the regime where shrinkage is most informative, and SURE adapts its strength without held-out data. Second, equal variance decouples the subproblems into $d$ independent closed-form regressions, eliminating the need for gradient descent altogether. The contrast with GBNSL is a bias-variance tradeoff. GBNSL's OLS estimator is asymptotically unbiased but inherits the variance of the sample covariance and precision estimates, which need enough samples relative to the dimension to stabilize, whereas SURE-Ridge's adaptive ridge trades a small bias for a large variance reduction. The contrast with NOTEARS and DAGMA is computational. Both pay a fixed gradient-step budget regardless of problem difficulty, optimize a non-convex objective vulnerable to finite-sample noise in the small-$n$ regime, and depend on a learning rate, a sparsity weight, and a continuation or augmented-Lagrangian schedule, none of which can be tuned against ground truth in the causal-discovery setting. SURE-Ridge therefore offers the best accuracy-compute-tunability trade-off in the small-sample regime targeted by this work.

\section{Conclusions}
\label{sec:conclusion}
This work proposes SURE-Ridge, a non-iterative, closed-form estimator for causal discovery in equal-variance linear Gaussian SEMs that combines $d$ node-wise ridge regressions with SURE-tuned regularization and an adaptive thresholding procedure for DAG extraction, requiring no per-problem hyperparameter tuning. On synthetic Gaussian SEMs, SURE-Ridge achieves the lowest SHD in the sparse, sample-limited regime against NOTEARS, DAGMA, and GBNSL baselines, and has a substantially faster run time than all three across every sample size tested, offering an attractive accuracy–compute tradeoff for constrained applications. Future work will assess the sample-complexity and convergence of SURE-Ridge.

\bibliographystyle{IEEEtran}
\bibliography{references}

\end{document}